# An Efficient Circle Detection Scheme in Digital Images Using Ant System Algorithm

Kaushik Chattopadhyay[1], Joydeep Basu[2], Amit Konar[3]
Department of Electronics and Telecommunication Engineering, Jadavpur University, Kolkata-700032, India
E-mails: [1]kaushik_chattt@yahoo.co.in, [2]joydeep_etce@yahoo.co.in, [3]konaramit@yahoo.co.in

**Abstract**
*Detection of geometric features in digital images is an important exercise in image analysis and computer vision. The Hough Transform techniques for detection of circles require a huge memory space for data processing hence requiring a lot of time in computing the locations of the data space, writing to and searching through the memory space. In this paper we propose a novel and efficient scheme for detecting circles in edge-detected grayscale digital images. We use Ant-system algorithm for this purpose which has not yet found much application in this field. The main feature of this scheme is that it can detect both intersecting as well as non-intersecting circles with a time efficiency that makes it useful in real time applications. We build up an ant system of new type which finds out closed loops in the image and then tests them for circles.*
**Key Words-**circle detection, edge detection, ant system algorithm, gray scale image, image processing.

## 1. Introduction

Detection of circles in digital images is a very important task from the view point of many applications in different domains like pattern recognition, machine vision etc. Presence of various shapes in digital images can be detected using Hough transform technique [1-6]. But a serious drawback of this technique is its huge time and space complexity. The time complexity for finding a shape using this technique grows exponentially with the number of parameters used to characterize the shape, e.g. for circles the number of parameters is 3, i.e. the co-ordinates of the centre and radius and this gives rise to a complexity of $O(n^3)$. Several papers have suggested methods to improve the technique [7]. In spite of that in many simple real time applications this method often turns out to be far from practical.

In this paper we propose a completely new, time and memory efficient, multi-agent searching scheme for detecting intersecting circles in an edge detected grayscale digital image. We consider the intersecting boundaries as a graph consisting of nodes and branches altogether. All the nodes and branches are found out and the incidence matrix of the graph is constructed. Using this information the closed loops in the graph are found out which are then tested for circles.

In Section 2 we define various terms. In Section 3 the agents that conduct the search, viz. ants are described. Section 4 describes how to arrange the collected data using various data structures so that data can be accessed very easily. The pseudo code of the scheme is also provided here. Section 5 describes how to test the detected loops for circles. In Section 6 we show the results of application of our method on some original sample images.

## 2. Branches and Nodes

In an edge detected image, a string of edge pixels stand for the boundary of an object present in the image. When we are talking about intersecting shapes, it is clear that such boundaries will cross each other. In a way, a **graph** (in not too technical a sense) is formed. A string of edge pixels, where each pixel is connected to only two adjacent pixels, is called a **branch**. A pixel where more than two branches meet is called a **node**. A branch is completely described by the pixel co-ordinates of that branch, taken in order. A node is completely specified if we have its co-ordinates along with the information that, to which other nodes it is connected and by which branches. This connectivity information is gathered in a matrix form which is formally called the ***incidence matrix***.

## 3. The 'Ants'

The search is done by one-pixel-agents that we shall call 'ants'. They occupy a single pixel and move one pixel at one step. An ant that moves, we call an **active** ant. We make them move over all the edge pixels according to some rules and extract information about the graph formed, i.e. they explore the edge image.

Each ant has its own co-ordinates which indicate on which pixel it is, and those co-ordinates are continuously updated as the ant moves. Before making a move an ant finds the number of directions available to move, by counting the number of edge pixels around the pixel it is resting upon, excluding the pixel it has just come from (this pixel is not to be covered again, i.e. the ant can not move back). For a branch, this number is one. For a node, it is at least two, and when an ant finds such a pixel, we say that it has discovered a node.

If an ant moves into a pixel on a branch, then it updates its own co-ordinates and stores the co-ordinates of the previous pixel in an array of its own called the *path[]* array. The figure below shows all possible adjacent pixels of *(x, y)*. If an ant moves from pixel *(x, y)* to the right upper corner pixel, then *(x, y)* is updated to *(x+1, y-1)* and similarly for other adjacent pixels.

| (x-1, y-1) | (x, y-1) | (x+1, y-1) |
|---|---|---|
| (x-1, y)   | (x, y)   | (x+1, y)   |
| (x-1, y+1) | (x, y+1) | (x+1, y+1) |

Figure 1. Co-ordinates of adjacent pixels

If an ant discovers a node, the co-ordinates of the node are recorded. The ant also discovers that the node it has come from and the new node are connected. This ant now becomes **inactive**, i.e. it ceases to move. New active ants are placed at the starting pixels of the branches that originate from this node. They continue further exploration of the graph.

In fact, the process of searching starts with placing an ant (the 'mother' ant) at a randomly selected edge pixel. This ant moves up to a node (this is the first discovered node) and generates new ants which travel along branches and create newer ants upon discovering new nodes. In this way a whole system of ants is build up.

So far we have observed three properties of the ants:

*Property 1:* All ants move at the same speed, i.e. one pixel at one step.

*Property 2:* Ants are generated at nodes.

*Property 3:* An ant remembers the node it has come from.

The above listed properties of ants have an interesting consequence. Suppose, between two nodes A and B, there is a shorter path S and a longer path L. The ants are generated at A and move towards B. The one along S will reach B sooner and place a new ant on the pixel that connects L to B, oblivious of the fact that there is already one coming towards B along L. The new ant will move towards A along L. Thus the two ants move toward each other and after some time their co-ordinates will become the same. We say that the two ants '**meet**'. Whenever two ants meet, we take advantage of that as follows:

**I:** Each of them had partially discovered the same branch. They recorded pixel co-ordinates of the branch from two sides in their *path [ ]* arrays. By joining the *path[ ]* arrays of these two ants back to back, we get an array which has all the pixel co-ordinates of the branch, in order.

**II:** Since each ant remembers from which node it comes from, when they meet, the connection between the nodes is discovered. We use this information to build up the incidence matrix.

After meeting each other, both the ants are made inactive. This prevents one ant from going through the same pixels that have already been visited by some other ant, which guarantees that each edge pixel is visited only once and by only one ant. We thus get another property:

*Property 4:* An ant ends up either reaching a node or meeting another ant and becomes inactive.

The scheme begins with a single active ant. Gradually their number rises as new nodes are discovered. But all the ants end up sometime as indicated by the above property, i.e. the number of active ants will become zero sometime and by that time not a single edge pixel remains to be discovered i.e., the graph is completely explored. In fact, we keep track of the number of active ants to know whether the exploration is complete or not.

## 4. Implementation

**4.1. Ants:** Each ant is represented as a structure with the following fields:
  **(x, y)**    its current co-ordinates

  *(xn, yn)*    co-ordinates of the node it has come from

  *path[ ]*     in this array it stores the pixel co-ordinates of the branch it is moving along

The ants are represented as individual blocks of a linked list. New blocks are added to the list with the generation of new ants. The block corresponding to a particular ant is deleted when it becomes inactive.

**4.2 Branches:** Each branch is represented as a structure which has the following fields:
  *index*    if index is *n*, we call that branch the *n*th branch or branch *n*

***branch_arr [ ]*** this array stores the co-ordinates of the pixels on the branch

Each branch is a block of a linked list. When a new branch is discovered, a block is added to the list. When an ant travels from one node to the other, it discovers a branch completely. Its ***path[ ]*** array becomes the ***branch_arr[ ]*** of the newly discovered branch. When two ants move toward each other from the terminal nodes of a branch and eventually meet, each has a part of the branch stored in their ***path[ ]*** arrays. We join these two ***path[ ]*** arrays back to back (to keep the order of the pixels right) and this becomes the ***branch_arr[ ]*** of this newly discovered branch. There is one variable called ***B***. This is initially set to zero and is incremented each time a branch is discovered and becomes the ***index*** of the newly discovered branch. After the ants finish their exploration, the variable ***B*** will be the total number of branches.

**4.3. Nodes:** Each node is also represented as a structure with the following fields:

    ***index***    if the index is ***n***, we call it the ***n***th node or node ***n***

    ***(x, y)***    co-ordinates of the node

    ***incidence [ ]***  an array

Several nodes are blocks of a linked list. As a new node is discovered, a new block is added to the list.

The array ***incidence[ ]*** has some special importance. If an ant coming from the ***i***th node reaches the ***j***th node or meets another ant coming from ***j***th node, through the ***k***th branch (i.e. branch with ***index k***), then we do the following operations:

    ***node i . incidence[k]*** = 1;

    ***node j . incidence[k]*** = - 1;

Otherwise the entries of these arrays are set to zero.
After the ants finish their exploration, the variable ***B*** will be the total number of branches and the index of the branch which is discovered last. Since branch indices are array positions of the ***incidence[ ]*** arrays, after exploration is complete, these arrays will have length that corresponds to the maximum branch index, i.e. ***B***. Let the total number of nodes be ***N***. Now we stack the ***incidence[ ]*** arrays of the nodes one below the other, starting with ***node 1*** at the top, ***node 2*** below that, then ***node 3*** and so on. Then it is clear that the ***NxB*** matrix so formed will be the ***incidence matrix*** of the graph.

**4.4. Pseudo Code**: First an ant, i.e. the mother ant is placed at any randomly selected edge pixel. It moves along the edge pixels and if it comes back to the starting pixel without encountering any pixel which leads to more than one direction, i.e. any node, then it has discovered a closed loop. The ***path[ ]*** array of this ant stores the co-ordinates of the pixels forming this loop. This array is tested for circularity.

But, if the mother ant reaches a node, it places new ants at the starting pixels that originate from this node, the node is recorded as the first node discovered and this ant itself becomes inactive. The pseudo code describing the movement of the ants from this point onwards is given.

Let the total number of active ants be ***N*** and the number of branches discovered be ***B***, at any time. Searching ends when ***N*** becomes zero, i.e. no active ants are left.

***while N*>0**
  ***i*=1;**
  ***while i<=N***
    find ***d***, the number of edge pixels available to move into around the pixel occupied by the ***i***th ant
    **if *d*>1**
      add a block to the list of nodes and record this pixel as a node
      ***B=B*+1;**
      add a block to the list of branches and make the ***path[ ]*** array of this ant the ***branch_arr[ ]*** of the new branch and the index of this branch is ***B***
      **if** the ant was coming from ***node p*** and has reached ***node q***
      **then *node p.incidence[B]* = +1;**
        ***node q.incidence[B]* = -1;**
      ***end***
      delete the block of this ***i***th ant from the list of ants
      place ***d*** new ants at the starting pixels of the ***d*** new branches originating from this node by adding ***d*** new blocks to the list of ants
      ***N=N+d-1;***
    **else if *d*=1**
      **if** the only pixel available to move into is occupied by some other ant //i.e. two ants meet//
      **then *B=B*+1;**
        add a new block to the list of branches
        join the ***path[ ]*** arrays of the ants back to back and make this the ***banch_arr[ ]*** of the new branch , the index of the branch is ***B***
        delete the block of the two ants from the list of ants
        ***N=N-2***
        **if** one ant was from ***node p*** and the other from ***node q***
        **then *node p.incidence[B]*=+1;**
          ***node q.incidence[B]*=-1;**
      ***end***
    ***else*** update the co-ordinates of the ***i***th ant according to the adjacent pixel it moves into
      record the co-ordinates of this pixel in the ***path[ ]*** array of the ***i***th ant
    ***end***
  ***end***

*i=i*+1
 *end*
*end*

## 5. Finding Circles

Once it is found from the incidence matrix which branches form a closed loop, we can easily find out if it represents a circular shape. If some branches (taken in order, of course) form a closed loop, then the **branch_arr[ ]** arrays of the branches (they are also taken in the same order as the branches) have the co-ordinates of the edge pixels that make the loop. We can analyze these co-ordinates whether they lie on a circle. We take three equidistant points on the loop at random and determine the equation of circle passing through them. Then the distance of each point on the loop is found from the centre of the circle and compared with the radius. Even if the loop looks perfectly like a circle, all points on it will not be at a same distance from the centre because of the error introduced inevitably due to quantization. So we have to make room for some tolerance which will depend on the radius of the circle. If some point fails to be within the range radius ± tolerance from the centre, then the loop is discarded from being a circle.

## 6. Simulation Results

The algorithm has been tested extensively on different images consisting of circular and non-circular objects. Figures below show two original images of different intricacies, having intersecting shapes and the circles extracted from them.

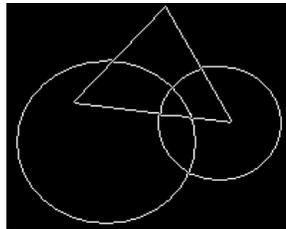
Fig.2.a Original image no. 1

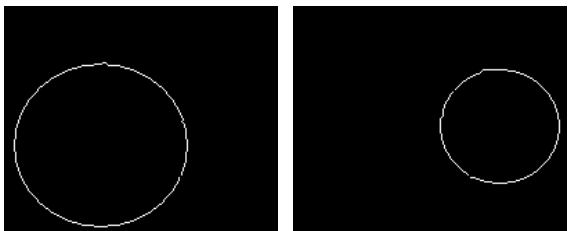
Fig. 2.b & 2.c Detected circles

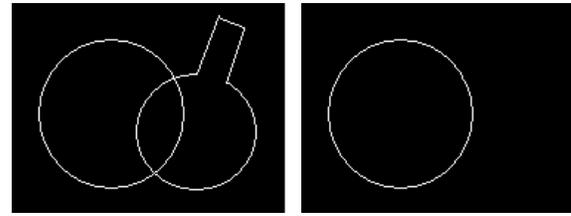
Fig. 3.a Original image no.2    Fig. 3.b Detected circle

## 7. Discussion and Conclusion

The first step of this scheme is the searching. As told before, each edge pixel is visited only once and by only one ant. Let the number of edge pixels be **$N_e$**. This can be taken very much greater than the number of nodes and branches, in most cases. Then the complexity of searching is **$O(N_e)$**. The next step is to find closed loops from the incidence matrix. This may take considerable time if the number of nodes and branches are too many. However, for various practical purposes where nodes and branches are less in number, this scheme is highly time efficient. Also notable is the memory efficiency of the scheme. The main memory usage goes into storing co-ordinates of edge pixels only.

There is no particular inclination towards circles. Once some closed loop is found, it can be tested for any shapes, like ellipses. However a limitation is imposed in this respect by the distortion due to digitization.